\documentclass[letterpaper, 10 pt, conference]{ieeeconf}  

\IEEEoverridecommandlockouts                              

\makeatletter
\let\NAT@parse\undefined
\makeatother

\usepackage[comma,numbers, compress]{natbib}

\usepackage[hidelinks]{hyperref}
\usepackage{url}
\usepackage{blindtext}
\usepackage{graphicx}
\usepackage[usenames,dvipsnames]{xcolor}
\usepackage[draft]{fixme}
\fxsetup{theme=color}
\usepackage{soul}

\usepackage{amsmath}
\usepackage{amssymb}

\usepackage{booktabs}
\usepackage{textcomp}
\usepackage{threeparttable}

\usepackage[capitalize]{cleveref}

\usepackage{tikz}
\usetikzlibrary{arrows}
\usetikzlibrary{positioning,calc}
\usetikzlibrary{decorations.pathreplacing}
\usetikzlibrary{decorations.markings}
\usetikzlibrary{fit}
\usetikzlibrary{shapes.callouts}
\usetikzlibrary{shapes.geometric}
\usetikzlibrary{matrix}

\usepackage[firstpage=true]{background}

\newcommand\copyrighttext{%
        \parbox{\textwidth}{
                \footnotesize
                In Proceedings of IEEE International Conference on Robotics and Automation (ICRA), Brisbane, Australia, May 2018, DOI: 10.1109/ICRA.2018.8461054 
        }
}

\SetBgContents{\copyrighttext}
\SetBgScale{1}
\SetBgColor{black}
\SetBgAngle{0}
\SetBgOpacity{1}
\SetBgPosition{current page.north}
\SetBgVshift{-0.8cm}

\graphicspath{{figures/}}
\title{\LARGE \bf
Planning Hybrid Driving-Stepping Locomotion \\ on Multiple Levels of Abstraction
}

\author{Tobias Klamt and Sven Behnke
\thanks{All authors are with Rheinische Friedrich-Wilhelms-Universit\"at Bonn, Computer Science Institute VI, 
		Autonomous Intelligent Systems, Endenicher Allee 19A, 53115 Bonn, Germany
        {\tt\small klamt@ais.uni-bonn.de, behnke@cs.uni-bonn.de}. This work was supported by the European Union's Horizon 2020 Programme under 
        Grant Agreement 644839 (CENTAURO).}%
}

\begin{document}

\maketitle
\thispagestyle{empty}
\pagestyle{empty}

\begin{abstract}

Navigating in search and rescue environments is challenging, since a variety of terrains has to be considered. Hybrid driving-stepping locomotion, as provided by our robot Momaro, is a promising approach. Similar to other locomotion methods, it incorporates many degrees of freedom---offering high flexibility but making planning computationally expensive for larger environments.

We propose a navigation planning method, which unifies different levels of representation in a single planner. In the vicinity of the robot, it provides plans with a fine resolution and a high robot state dimensionality. With increasing distance from the robot, plans become coarser and the robot state dimensionality decreases. We compensate this loss of information by enriching coarser representations with additional semantics. Experiments show that the proposed planner provides plans for large, challenging scenarios in feasible time.

\end{abstract}


\section{Introduction}

Hybrid driving-stepping locomotion is a flexible approach to traverse many types of terrain since it combines the advantages of both, wheeled and legged, locomotion types. However, due to its high robot state dimensionality, planning respective paths is challenging.

In our previous work~\cite{klamtanytime} we presented an approach to plan hybrid driving-stepping locomotion paths for our robot Momaro~\cite{Schwarz:ICRA2016} even for very challenging terrain such as staircases with additional obstacles on it. The planner prefers omnidirectional driving whenever possible and considers individual steps in situations where driving is not possible. The individual configuration of ground contact points (robot footprint) is considered at any time. During planning, steps are represented as abstract manoeuvres which are expanded to detailed motion sequences before executing them. For small scenarios, this method generates high quality paths in feasible time with bounded suboptimality. Due to the high dimensionality of the robot configuration, the explored state space increases rapidly for larger scenarios and makes planning expensive. This effect is not unique for hybrid driving-stepping locomotion but affects high-dimensional planning in many applications such as locomotion planning for robots with tracked flippers or manipulation planning.

The search space can be reduced by choosing a coarser resolution or describing the robot and its manoeuvres in a more abstract way with less degrees of freedom (DoF). However, a fine resolution is key to navigate the robot precisely through challenging terrain. Moreover, only using a more abstract robot description is difficult, since the planning result shall be a path which can be executed by the robot with its given number of DoF.

Coarse-to-fine planning approaches~\cite{bohlin2001path, kohrt2012cell} address this problem by generating a rough plan first and refine the resulting path to the desired resolution and number of DoF in a second step. Especially in challenging, cluttered terrain, this procedure bears the risk of only finding expensive paths due to the lack of detail in the initial search.  

\begin{figure}
\centering
\input{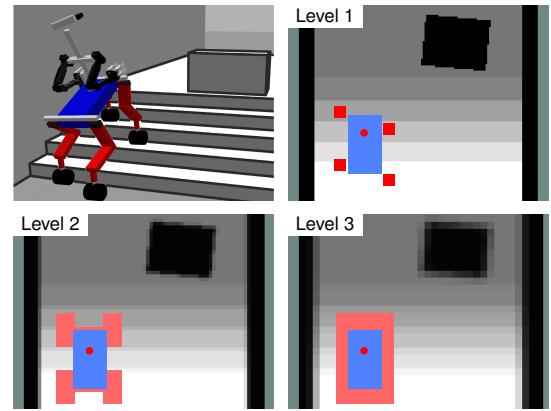}
\vspace{-1.2cm}
\caption{Momaro on a staircase, visualized in three representation levels. Maps show terrain heights (olive = unknown).}
\label{fig:teaser_picture}
\end{figure}

We present a method which plans hybrid driving-stepping locomotion on three different levels of representation (see~\cref{fig:teaser_picture}). In the vicinity of the robot, a representation with a high resolution and a high number of DoF is used to find paths which can be executed by the robot. With increasing distance from the robot, the resolution gets coarser and the robot is described with less DoF. These path segments are situated further in the future which comes along with a higher degree of uncertainty and less accurate sensor information. We compensate this loss of information for higher levels of representation by enriching the representation with additional semantics. All levels of representation are unified in a single planner. We further present methods to refine path segments into more detailed levels of representation. This decreases the number of necessary replanning steps. Replanning is only initiated if costs indicate that a situation is wrongly assessed in the coarser representation. In addition, we introduce a heuristic, based on the most abstract level of representation.

Experiments show that, compared to our previous work, this approach can handle much larger scenarios in feasible planning time while the path quality stays comparable.


\section{Related Work}

Multiple works addressed path planning for challenging environments, either by driving~\cite{ziaei2014global, howard2007optimal, brunner2012motion} or walking with quadruped robots~\cite{wermelinger2016navigation, perrin2016continuous}. To our knowledge, there exist no approaches for hybrid driving-stepping path planning in challenging terrain, except our previous work~\cite{klamtanytime}.

A common idea to accelerate planning for larger scenarios is the usage of multiresolutional approaches. Behnke~\cite{behnke2003local} proposed a general concept for A*-based multiresolution planning with a decreasing resolution with increasing distance from the robot. Gonz\'{a}lez-Sieira et al.~\cite{gonzalez2016adaptive} apply high resolution in areas of high environment complexity. Resolution decreases with increasing distance from these areas. Similarly, Pivtoraiko et al.~\cite{pivtoraiko2008differentially} apply different sets of state transitions to different areas of the environment. Bohlin~\cite{bohlin2001path} generates an initial plan in a coarse resolution first and refines this plan into a finer resolution. Since high resolution planning is only applied to parts of the map, the search space decreases and planning performance increases, compared to pure high resolution planning. One of the main challenges in multiresolutional approaches is the definition of feasible transitions between the different resolutions. All of the presented approaches face the problem that a coarse resolution representation neglects information and thus is not capable of representing challenging terrain features in sufficient detail---which might lead to wrong or bad plans. 

Planning for systems with high-dimensional motion flexibility quickly reaches its limits for larger environments since the search space grows exponentially. Similar to multiresolution planning, several approaches utilize multiple representations with different planning dimensionalities to decrease planning complexity. Kohrt et al.~\cite{kohrt2012cell} generate an initial plan in a low-dimensional search space and replan in the high-dimensional search space by only considering those states that are part of the low-dimensional plan. Gochev et al.~\cite{gochev2011path} plan a path in a low-dimensional search space and only switch to high-dimensional planning in those areas where low-dimensional planning cannot find a solution. Similarly, Zhang et al.~\cite{zhang2012combining} plan in 2D and switch to high-dimensional planning in the robot vicinity and at key points. As described for multiresolution planning, planning with multiple robot configuration dimensionalities might lead to wrong or bad plans, since a low-dimensional robot representation might assess challenging situations wrongly.

To achieve further planning acceleration, it is an obvious idea to combine multiresolution and multidimensional planning. However, only few works, such as by Petereit et al.~\cite{petereit2013mobile} address this. Different planning dimensionalities and resolutions are applied by using different sets of motion primitives. A fine resolution is only considered close to the start and goal pose and close to obstacles. A high planning dimensionality is considered for states which will be reached within a given time interval. This allows the planner to provide detailed plans close to the robot while planning times stay feasible. The drawbacks of both, multiresolutional and multidimensional planning also apply to this work.

The platforms which are used in the presented works are quite limited in their configuration capabilities, compared to our robot Momaro. Our approach applies multiresolution and multidimensional planning to the challenging problem of hybrid driving-stepping locomotion. Furthermore, we compensate the loss of information for coarser resolutions and low-dimensional robot representations by enriching those representations with additional semantic features.


\section{Hardware}

\begin{figure}
\centering
\includegraphics[width=.8 \linewidth]{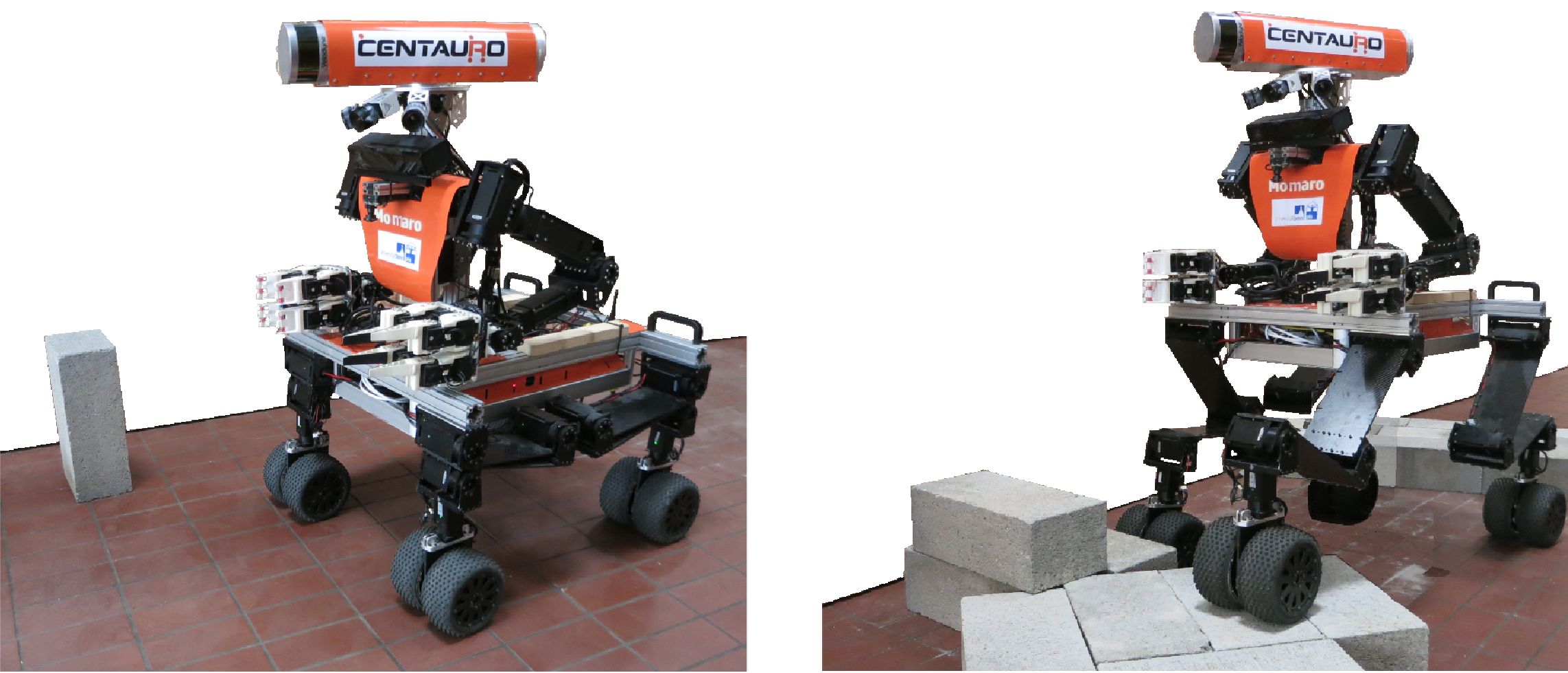}
\caption{Our wheeled-legged robot Momaro is capable of omnidirectional driving (left) and stepping (right).}
\label{fig:momaro}
\end{figure}

We use our mobile manipulation robot Momaro~\cite{Schwarz:ICRA2016} (see \cref{fig:momaro}). It offers omnidirectional driving through its four articulated legs ending in directly driven 360\textdegree~steerable pairs of wheels. The unique design enables manoeuvres which are neither realizable by pure driving nor pure walking robots such as shifting a single foot while maintaining ground contact and thus changing the robot footprint under load. Active leg movements are restricted to the sagittal plane since each leg consists of three pitch joints.

Sensor inputs come from an IMU and a continuously rotating Velodyne Puck 3D laser scanner at the robot head which provides a spherical field-of-view. The laser-range measurements are registered and aggregated to a 3D environment map using the method of Droeschel et al.~\cite{droeschel2017continuous}.


\section{Approach}

\begin{figure*}
\centering
\begin{tikzpicture}[
 	font=\sffamily\footnotesize,
    every node/.append style={text depth=.2ex},
	l/.style={font=\sffamily\scriptsize},
]


\definecolor {mygreen}{RGB}{120, 200, 120};
\definecolor {myblue}{RGB}{100, 150, 255};
\definecolor {myred}{RGB}{240, 110, 110};
\definecolor {robot_base}{RGB}{80, 130, 255};

\fill[myred] (0.0,3.0) rectangle ++(1,1.2);
\fill [myblue] (0.0,1.8) rectangle ++(1,1.2);
\fill[mygreen] (0.0,0.6) rectangle ++(1,1.2);

\node[] at (0.52,4.4) {Level};
\node[] at (0.52,3.58) {1};
\node[] at (0.52,2.38) {2};
\node[] at (0.52,1.18) {3};

\node[] at (2.75,4.4) {Map Resolution};
\node[] at (2.5,3.9) [l, anchor=west] {$\bullet$ 2.5\,cm};
\node[] at (2.5,3.6) [l, anchor=west] {$\bullet$ 64 orient.};
\node[] at (2.5,2.7) [l, anchor=west] {$\bullet$ 5.0\,cm};
\node[] at (2.5,2.4) [l, anchor=west] {$\bullet$ 32 orient.};
\node[] at (2.5,1.5) [l, anchor=west] {$\bullet$ 10\,cm};
\node[] at (2.5,1.2) [l, anchor=west] {$\bullet$ 16 orient.};

\node[] at (6.52,4.4) {Map Features};
\node[] at (6,3.9) [l, anchor=west] {$\bullet$ Height};
\node[] at (6,2.7) [l, anchor=west] {$\bullet$ Height};
\node[] at (6,2.4) [l, anchor=west] {$\bullet$ Height Difference};
\node[] at (6,1.5) [l, anchor=west] {$\bullet$ Height};
\node[] at (6,1.2) [l, anchor=west] {$\bullet$ Height Difference};
\node[] at (6,0.9) [l, anchor=west] {$\bullet$ Terrain Class};

\node[] at (10.75,4.4) {Robot Representation};
\fill[robot_base] (11,3.3) rectangle ++(1,0.6);
\fill[red] (10.9,3.05) rectangle ++(0.2,0.2);
\fill[red] (12.15,3.05) rectangle ++(0.2,0.2);
\fill[red] (11.1,3.95) rectangle ++(0.2,0.2);
\fill[red] (11.9,3.95) rectangle ++(0.2,0.2);
\fill[red] (11.7,3.6) circle (0.07);
\draw[-latex](11.1,3.15) -- ++(0.25, 0);
\draw[-latex](10.9,3.15) -- ++(-0.25, 0);
\draw[-latex](12.35,3.15) -- ++(0.25, 0);
\draw[-latex](12.15,3.15) -- ++(-0.25, 0);
\draw[-latex](11.3,4.05) -- ++(0.25, 0);
\draw[-latex](11.1,4.05) -- ++(-0.25, 0);
\draw[-latex](12.1,4.05) -- ++(0.25, 0);
\draw[-latex](11.9,4.05) -- ++(-0.25, 0);
\draw[-latex](11.77,3.6) -- ++(0.25, 0);
\draw[-latex](11.7,3.67) -- ++(0,0.25);
\draw[-latex](11.63,3.6) -- ++(-0.25,0);
\draw[-latex](11.7,3.53) -- ++(0,-0.25);
\draw[-latex](12.2,3.625) arc (5:60:0.35);
\draw[-latex](12.2,3.575) arc (-5:-60:0.35);

\fill[red!60] (10.9,1.85) rectangle ++(0.45,0.3);
\fill[red!60] (10.9,2.65) rectangle ++(0.45,0.3);
\fill[red!60] (12.1,1.85) rectangle ++(0.45,0.3);
\fill[red!60] (12.1,2.65) rectangle ++(0.45,0.3);
\fill[red!60] (12.275,1.85) rectangle ++(0.1,1.0);
\fill[robot_base] (11,2.1) rectangle ++(1,0.6);
\fill[red] (11.7,2.4) circle (0.07);
\draw[-latex](11.35,1.95) -- ++(0.25, 0);
\draw[-latex](10.9,1.95) -- ++(-0.25, 0);
\draw[-latex](12.55,1.95) -- ++(0.25, 0);
\draw[-latex](12.1,1.95) -- ++(-0.25, 0);
\draw[-latex](11.77,2.4) -- ++(0.25, 0);
\draw[-latex](11.7,2.47) -- ++(0,0.25);
\draw[-latex](11.63,2.4) -- ++(-0.25,0);
\draw[-latex](11.7,2.33) -- ++(0,-0.25);
\draw[-latex](12.2,2.425) arc (5:60:0.35);
\draw[-latex](12.2,2.375) arc (-5:-60:0.35);

\fill[red!60] (10.7,0.65) rectangle ++(1.6,1.1);
\fill[robot_base] (11,0.9) rectangle ++(1,0.6);
\fill[red] (11.7,1.2) circle (0.07);
\draw[-latex](11.77,1.2) -- ++(0.25, 0);
\draw[-latex](11.7,1.27) -- ++(0,0.25);
\draw[-latex](11.63,1.2) -- ++(-0.25,0);
\draw[-latex](11.7,1.13) -- ++(0,-0.25);
\draw[-latex](12.2,1.225) arc (5:60:0.35);
\draw[-latex](12.2,1.15) arc (-5:-60:0.35);

\node[] at (14.92,4.4) {Action Semantics};
\node[] at (14.5,3.9) [l, anchor=west] {$\bullet$ Individual};
\node[] at (14.73,3.6) [l, anchor=west] {Foot Actions};
\node[] at (14.5,2.7) [l, anchor=west] {$\bullet$ Foot Pair};
\node[] at (14.73,2.4) [l, anchor=west] {Actions};
\node[] at (14.5,1.5) [l, anchor=west] {$\bullet$ Whole Robot};
\node[] at (14.73,1.2) [l, anchor=west] {Actions};

\draw[](0,0.6) rectangle (16.5,4.6);
\draw[](1,0.6) -- ++(0,4);
\draw[](4.5,0.6) -- ++(0,4);
\draw[](8.5,0.6) -- ++(0,4);
\draw[](13.0,0.6) -- ++(0,4);
\draw[](0,4.2) -- ++(16.5,0);
\draw[](0,3.0) -- ++(16.5,0);
\draw[](0,1.8) -- ++(16.5,0);

\draw[thick](1.8,0.7) -- ++(0.6,3.3) -- ++(-1.2,0) -- ++(0.6,-3.3);
\draw[thick](9.3,0.7) -- ++(0.6,3.3) -- ++(-1.2,0) -- ++(0.6,-3.3);
\draw[thick](4.65,0.7) -- ++(1.2,0) -- ++(-0.6,3.3) -- ++(-0.6,-3.3);
\draw[thick](13.2,0.7) -- ++(1.2,0) -- ++(-0.6,3.3) -- ++(-0.6,-3.3);

\end{tikzpicture}
\caption{The planner includes three levels of representation with decreasing resolution and robot configuration dimensionality. To compensate the loss of information, the semantics for both the environment representation and the robot actions increase.}
\label{fig:representation_level_concept}
\end{figure*}

Input to our method is a height map with a resolution of 2.5\,cm which is generated from the 3D environment map. In the vicinity of the robot, height information is very precise. With increasing distance from the robot, the accuracy decreases due to measurement errors. Planning is done on foot and body costs. The ground contact costs $C_\text{GC}$ describe the costs to place an individual ground contact element (e.g., a foot or a foot pair) in a given configuration on the map. $C_\text{GC}$ includes information about the terrain surface and obstacles in the vicinity. The body costs $C_\text{B}(\vec{r}_b)$ describe the costs to place the robot base \mbox{$\vec{r}_b=(r_x, r_y, r_\theta)$} with its center position $r_x, r_y$ and its orientation $r_\theta$ on the map. $C_B(\vec{r}_b)$ include information about obstacles under the robot base and about the terrain slope under the robot. The generation of $C_\text{GC}$ and $C_\text{B}$ from the height map varies between the different levels of representation and may contain several steps. Ground contact costs and body costs are combined to pose costs $C(\vec{r})$ which describe the costs to place the robot in a given configuration $\vec{r}$ on the map. 

Path planning is realized through an A*-search with anytime characteristics (ARA*~\cite{likhachev2003ara}) on pose costs. For the current search pose, feasible neighbour poses are generated during the search. They can either be reached by omnidirectional driving or by stepping-related motions. Stepping-related motions are only considered in the vicinity of obstacles where driving is infeasible. Steps are described as abstract steps, the direct transition between a pre-step to an after-step pose. The detailed motion sequence for steps is not considered during planning but generated before execution.    

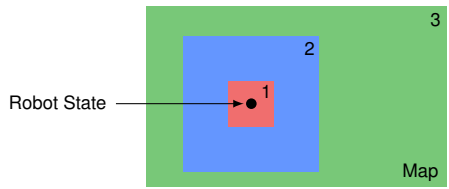
\begin{figure}
	\centering
	\begin{tikzpicture}[
 	font=\sffamily\footnotesize,
    every node/.append style={text depth=.2ex},
	l/.style={font=\sffamily\scriptsize},
]

\definecolor {mygreen}{RGB}{120, 200, 120};
\definecolor {myblue}{RGB}{100, 150, 255};
\definecolor {myred}{RGB}{240, 110, 110};

\fill [mygreen,inner sep = 0] (0.0,0.0) rectangle ++(4,2.4);
\fill[myblue,inner sep=0] (0.5,0.2) rectangle ++(1.8,1.8);
\fill[myred,inner sep=0] (1.1,0.8) rectangle ++(0.6,0.6);
\fill[black,inner sep=0] (1.4,1.1) circle (2pt);

\fill[white,inner sep=0] (-2.0,0.00) rectangle ++(2.0,0.2);
\fill[white,inner sep=0] (4,0.00) rectangle ++(2.0,0.2);

\node[l] at (1.6,1.25)  {1};
\node[l] at (2.175,1.825) {2};
\node[l] at (3.85,2.2) {3};
\node[l,inner sep=0] at (3.65,0.2) {Map};

\draw[-latex]   (-0.4,1.1) -- (1.3,1.1) node [pos=0, left,l] {Robot State};

\end{tikzpicture}
	\caption{Size and position of the different levels of representation. \emph{Level~1} covers the vicinity of the robot. \emph{Level~2} is also robot centered and medium sized. \emph{Level~3} covers the whole map.}
	\label{fig:level_size_concept}
\end{figure}

The environment and the robot are described in three different levels of representation with different sizes. In the vicinity of the robot, we use a fine resolution and a high robot configuration dimensionality for planning. We call this \emph{Level~1} representation. With increasing distance from the current robot position, the environment and the robot are represented on higher levels with a coarser resolution and a robot representation with lower dimensionality. This is reasonable, since those parts of the plan are reached in the further future and thus are more uncertain. Moreover, sensor measurements become less precise with increasing distance from the robot. At the same time, we compensate this loss of detail by enriching the environment representation with additional features, which increase the understanding of the situation. Pose costs and robot actions use these semantic features. Higher levels of representation can be derived from lower levels of representation. The approach is visualized in~\cref{fig:representation_level_concept}. Level sizes and positions are shown in~\cref{fig:level_size_concept}. For a planning task, the planner only performs a single planning run while including all three levels of representation. Hence, it is important that the same action carries the same costs in different levels of representation to make planning consistent over all levels. Moreover, the transition between the different levels of representation is challenging. All three levels of representation and the transition between them are described in detail in the following sections.

The resulting path consists of segments in multiple levels of representation. As described before, the contained steps are abstract manoeuvres. Abstract steps in the initial path segment are expanded to detailed motion sequences before executing them. Roll and pitch motions of the robot base as well as single foot shifts stabilize the robot to perform each step safely. In addition, foot heights are derived. See our previous work~\cite{klamtanytime} for more details. 

Steps are only expanded for path segments in \emph{Level~1} which is based on our previous work. For higher levels, representations are not detailed enough to derive concrete robot motions. As the robot executes the initial path segment, more measurements are made and a more detailed environment representation becomes available for path segments which have been represented in higher levels before. The path is updated with these updated representations. This can either be done by replanning the whole path or by transforming the respective path segments into more detailed representations, as described in~\cref{sec:continuous_refinement}. We call this coarse-to-fine transformation {\em refinement}.


\subsection{Representation Level 1}

\begin{figure}
\centering
\includegraphics[width=0.65\linewidth]{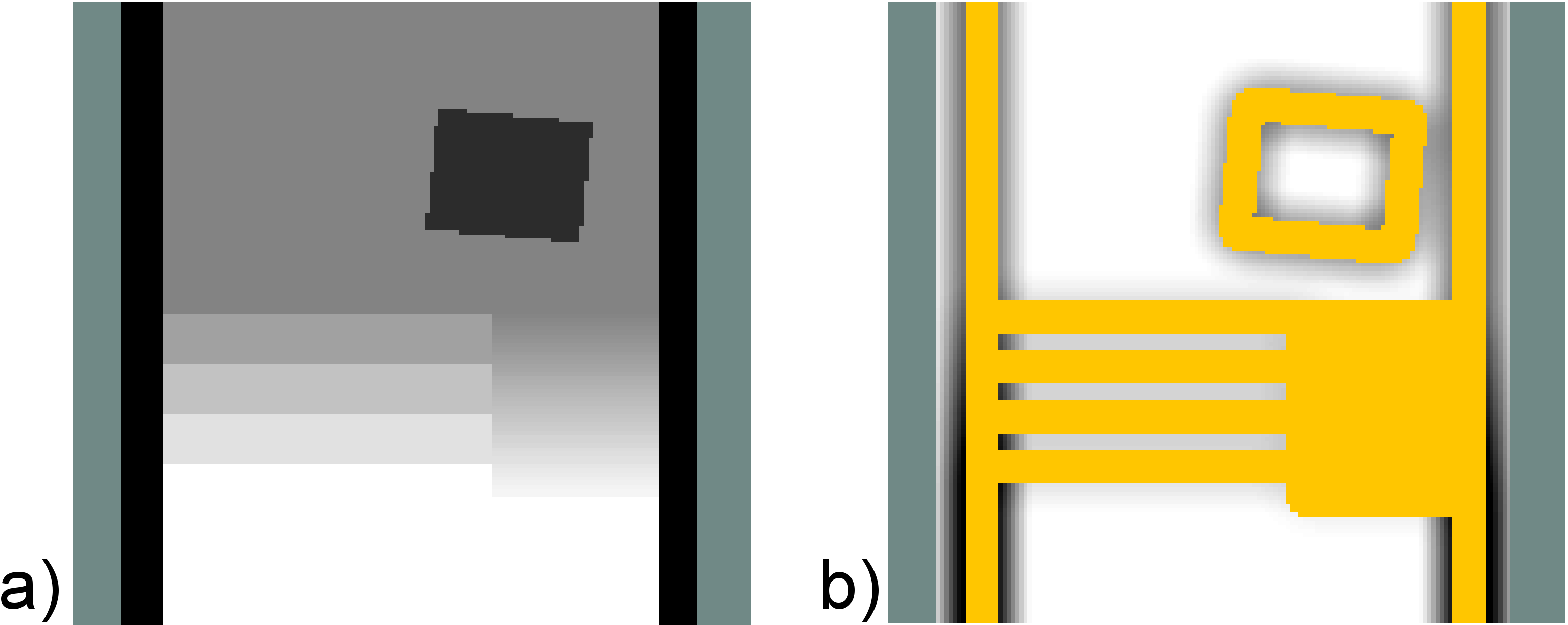}
\caption{a) \emph{Level~1} height map showing a corridor with a flight of stairs, an untraversable steep ramp and an obstacle, b) respective foot cost map (yellow = untraversable by driving, olive = unknown).}
\label{fig:level1_maps}
\end{figure}

\emph{Level~1} is based on the approach which we presented in our previous work~\cite{klamtanytime}. Input is a height map with a resolution of 2.5\,cm. We derive local unsigned height differences between neighbour cells from this height map to generate ground contact costs for each individual foot. Base costs are derived from the height map itself. A height map and the derived foot costs can be seen in~\cref{fig:level1_maps}. In this level of representation, a robot pose $\vec{r}=(\vec{r}_b, f_1, ..., f_4)$ is represented through the robot base configuration $\vec{r}_b$ and the individual longitudinal foot positions $f_1, ... , f_4$. At each position, the robot can have 64 different discrete orientations. 

\begin{figure}
\centering
\input{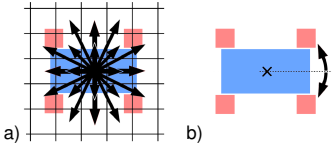}
\caption{Driving neighbour poses can be found by either a) straight moves with fixed orientation within a 20-position-neighbourhood or b) by turning on the spot to the next discrete orientation.}
\label{fig:driving_neighbours}
\end{figure}

\begin{figure}
\centering
\input{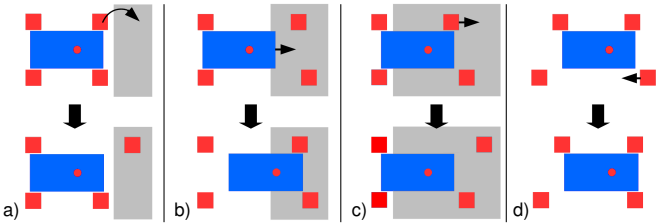}
\caption{\emph{Level~1} stepping-related manoeuvres: a) Abstract step, b) longitudinal base shift, c) shifting a front foot forward, and d) shifting any foot back to its neutral position.}
\label{fig:stepping_neighbours_level1}
\end{figure}

Feasible driving neighbour poses can be found within a 20-position-neighbourhood and by turning on the spot to the next discrete orientation, as shown in~\cref{fig:driving_neighbours}. If the robot is close to an obstacle, additional stepping-related manoeuvres are considered which are visualized in~\cref{fig:stepping_neighbours_level1}. Those can be a discrete step, a longitudinal base shift manoeuvre, shifting individual feet forward or shifting individual feet towards their neutral position. We define the neutral robot pose as the pose visualized in~\cref{fig:stepping_neighbours_level1}\,a,\,top. The costs for the presented manoeuvres are based on the foot and body costs, the individual robot elements induce during the manoeuvre. 

As an extension of the previous work, we want the robot to align its orientation with the stair orientation, when climbing those. This is desirable, since the kinematic only allows for leg movements in the sagittal plane and since this behavior can also be observed when humans climb stairs by themselves or teleoperate robots to do so. If, after a stepping manoeuvre, the two front/rear feet have the same longitudinal position but stand on different heights, this indicates that the robot is not aligned with the stairs it climbs. By punishing such a configuration with an additional cost term, we achieve the desired behavior.


\subsection{Representation Level 2}
\label{sec:lvl2}

\begin{figure}
\centering
\begin{tikzpicture}[
 	font=\sffamily\footnotesize,
    every node/.append style={text depth=.2ex},
	l/.style={font=\sffamily\scriptsize},
]

\draw[](6,0.0) -- ++(1.6,0);
\draw[](6,0.4) -- ++(1.6,0);
\draw[](6,0.8) -- ++(1.6,0);
\draw[](6,1.2) -- ++(1.6,0);
\draw[](6,1.6) -- ++(1.6,0);
\draw[](6,0.0) -- ++(0,1.6);
\draw[](6.4,0.0) -- ++(0,1.6);
\draw[](6.8,0.0) -- ++(0,1.6);
\draw[](7.2,0.0) -- ++(0,1.6);
\draw[](7.6,0.0) -- ++(0,1.6);

\node[l] at (6.2,0.2) {1};
\node[l] at (6.6,0.2) {3};
\node[l] at (7,0.2) {3};
\node[l] at (7.4,0.2) {1};
\node[l] at (6.2,0.6) {3};
\node[l] at (6.6,0.6) {9};
\node[l] at (7,0.6) {9};
\node[l] at (7.4,0.6) {3};
\node[l] at (6.2,1) {3};
\node[l] at (6.6,1) {9};
\node[l] at (7,1) {9};
\node[l] at (7.4,1) {3};
\node[l] at (6.2,1.4) {1};
\node[l] at (6.6,1.4) {3};
\node[l] at (7,1.4) {3};
\node[l] at (7.4,1.4) {1};

\node[] at (5.5,0.8) {$\frac{1}{64}$ x};

\draw[](1.8,0.0) -- ++(2.0,0);
\draw[](1.8,0.4) -- ++(2.0,0);
\draw[](1.8,0.8) -- ++(2.0,0);
\draw[](1.8,1.2) -- ++(2.0,0);
\draw[](1.8,1.6) -- ++(2.0,0);
\draw[](2,-0.2) -- ++(0,2.0);
\draw[](2.4,-0.2) -- ++(0,2.0);
\draw[](2.8,-0.2) -- ++(0,2.0);
\draw[](3.2,-0.2) -- ++(0,2.0);
\draw[](3.6,-0.2) -- ++(0,2.0);

\draw[blue, thick, dashed](1.98, -0.02) rectangle (3.62, 1.62);
\draw[red, thick](2.38, 0.38) rectangle (3.22, 1.22);

\node[l] at (1.6,-0.05) {a)};
\node[l] at (5.0,-0.05) {b)};

\end{tikzpicture}
\caption{Subsampling method: a)~For a \emph{Level~2} cell (red square) a 4$\times$4-window (blue square) of \emph{Level~1} cells is considered. b)~Normalized binomial distribution to weigh heights and height differences.}
\label{fig:subsampling}
\end{figure}
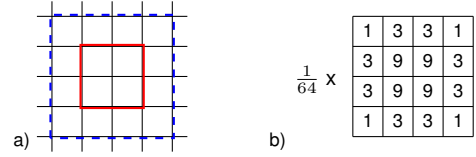

\begin{figure}
\centering
\input{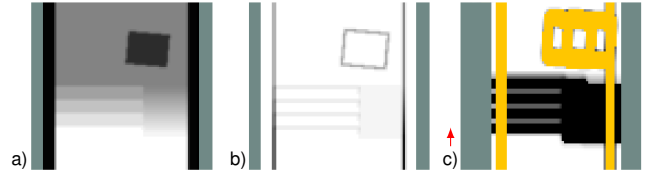}
\caption{\emph{Level~2}: a)~height map, b)~height difference map, c)~foot area pair cost map for the orientation indicated by the red arrow.}
\label{fig:level2_maps}
\end{figure}

We use the input height map with a resolution of 2.5\,cm to compute the \emph{Level~2} representation consisting of a height map and a height difference map with a resolution of 5\,cm (see~\cref{fig:level2_maps}\,a,b). According to the Nyquist-Shannon sampling theorem, subsampling has to come along with smoothing. To satisfy this theorem, we subsample the \emph{Level~1} height map as shown in~\cref{fig:subsampling}. Each \emph{Level~2} height value is computed from the normalized, weighted sum of a 4$\times$4-region of \emph{Level~1} height values. We use a binomial distribution for weighing. A \emph{Level~2} height difference map is generated in the same manner: We generate a \emph{Level~1} height difference map by computing local height differences on the \emph{Level~1} height map. This height difference map is then subsampled to a \emph{Level~2} height difference map.

To decrease the robot configuration space dimensionality, we accumulate individual feet to pairs. This is intuitive, since we observe a tendency to pairwise foot movement in \emph{Level~1} paths.  Moreover, instead of describing each foot position precisely, we use foot areas as a more abstract description. We know, that a foot will be placed somewhere in the respective area but since the representation contains some time-related and measurement-related imprecision, a knowledge of the accurate foot position is not necessary. A \emph{Level~2} robot pose $\vec{r}=(\vec{r}_b, f_\text{f}, f_\text{r})$ is consequently represented by its robot base pose $\vec{r}_b$ and its relative longitudinal front and rear foot area pair coordinates $f_\text{f}$ and $f_\text{r}$. Note that our platform and planner only allow sagittal leg movement. Lateral foot coordinates are fixed and thus a single variable is sufficient to describe each foot area pair.

We use the generated \emph{Level~2} representation to compute ground contact and body costs. Body cost computation is similar to \emph{Level~1} and only relies on height information. Ground contact costs 
\begin{equation}
	C_\text{GC,2}=1+k_\text{1} \cdot \bigtriangleup H_\text{avg} \text{,}
\end{equation}
where $k_\text{1}$ = 107, are costs to place foot area pairs on the map and are generated from the average height differences $\bigtriangleup H_\text{avg}$ in the respective area. A \emph{Level~2} foot area pair cost map can be seen in~\cref{fig:level2_maps}\,c. Again, a punishing cost term is introduced for after-step poses with different average heights under neighbouring foot areas.

\begin{figure}
\centering
\input{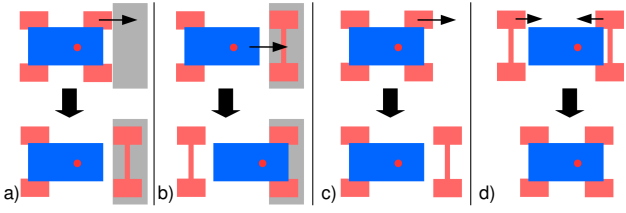}
\caption{\emph{Level~2} stepping-related manoeuvres: a)~Step, b)~longitudinal base shift, c)~move the front foot pair forward, and d) move any foot pair towards its neutral position.}
\label{fig:stepping_neighbours_level2}
\end{figure}

The robot actions are defined accordingly. Driving neighbours can be found similar to \emph{Level~1} but with a doubled action resolution of 5\,cm and 32 discrete robot orientations at each position. Additional stepping-related manoeuvres differ from \emph{Level~1} since the robot is only able to move foot pairs instead of individual feet. If the robot is close to an obstacle, it may step with a foot pair or perform another stepping related manoeuvre, as visualized in~\cref{fig:stepping_neighbours_level2}. To motivate stepping manoeuvres, we define a maximum height difference $\bigtriangleup H_\text{max,drive}$ for the foot area center coordinate which can be overcome by driving. Larger height differences only can be traversed by stepping.

The costs for such a foot pair manoeuvre are the concatenated costs for each individual foot action as described for \emph{Level~1}. If, for example, the robot steps with its front foot pair as visualized in~\cref{fig:stepping_neighbours_level2}\,a, the costs for this manoeuvre are the sum of the costs for a step with the front left foot and a step with the front right foot. Since \emph{Level~2} foot pair area costs differ from \emph{Level~1} foot costs, we reparametrized the manoeuvre cost computation. We do this by performing foot pair manoeuvres in a variety of basic scenarios (e.g., drive/turn on a patch of flat/rough underground, step up different height differences, do a base shift) in both representation levels and manually tune the \emph{Level~2} cost parameters until the costs for those manoeuvres in both levels vary by $\leq$\,5\%. 

\begin{figure}
\centering
\input{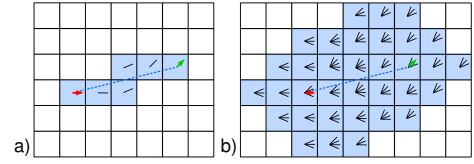}
\caption{Generating a set of feasible robot base poses for path refinement: a)~For a given start ($\vec{r}_\text{1,i}$, red arrow) and goal ($\vec{r}_\text{1,i+1}$, green arrow) robot base pose, we generate a set of feasible robot base poses (black lines) by interpolating between the two. b)~Inflation by two position steps and one orientation step.}
\label{fig:refining}
\end{figure}

During planning and execution, it is an important feature to refine \emph{Level~2} path segments into \emph{Level~1}. To refine a \emph{Level~2} path segment between two successive poses $\vec{r}_\text{2,i}$ and $\vec{r}_\text{2,i+1}$, we transform both poses into \emph{Level~1} and generate a set $S$ of feasible robot base poses by interpolating between $\vec{r}_\text{1,i}$ and $\vec{r}_\text{1,i+1}$. $S$ is then inflated with a radius of two position steps and one orientation step as visualized in~\cref{fig:refining}. A local planner, which is restricted to $S$, searches for a \emph{Level~1} path between $\vec{r}_\text{1,i}$ and $\vec{r}_\text{1,i+1}$. If
\begin{itemize}
	\item either one of the two poses becomes infeasible when transformed to \emph{Level~1} because \emph{Level~2} assessed the given situation wrongly or 
	\item the costs for the refined \emph{Level~1} path differs by \textgreater\,25\% from the original costs for the path segment, 
\end{itemize} 
we call this path segment {\em not refineable}.

\subsection{Representation Level 3}

We apply the described subsampling process (see~\cref{sec:lvl2}) to generate a \emph{Level~3} height map and height difference map with a resolution of 10\,cm from the \emph{Level~2} height map and height difference map. To increase the semantics of the environment representation, we categorize each \emph{Level~2} map cell into one of the following terrain classes:
\begin{itemize}
	\item \emph{flat}: easily traversable by driving,
	\item \emph{rough}: traversable by driving with high effort,
	\item \emph{step}: includes height differences which are too large to be traversed by driving but can be traversed by stepping,
	\item \emph{wall}: occurring height differences are too large to be traversed by stepping, and
	\item \emph{unknown}: cell cannot be classified.
\end{itemize}
First, we search for cells of the terrain type \emph{step}. This is done by searching for cell pairs $c_i$ and $c_j$ that fulfill the following criteria:
\begin{itemize}
	\item $\bigtriangleup H(c_i) < \bigtriangleup H_\text{max,drive}$: $c_i$ is on a drivable surface,
	\item $\bigtriangleup H(c_j) < \bigtriangleup H_\text{max,drive}$: $c_j$ is on a drivable surface,
	\item $\left\lVert c_\text{i} - c_j\right\rVert < 0.45$\,m: The distance between $c_i$ and $c_j$ is within a maximum step length, and
	\item for the set $T$ of cells $c_k$ on the straight line between $c_i$ and $c_j$, $C_\text{GC}(c_k) = \infty$ counts for all cells $c_k \in T$: A direct foot movement from $c_i$ to $c_j$ requires a step. 
\end{itemize}
For all pairs of $c_i$ and $c_j$ which fulfill these criteria, each cell $c_s \in c_i \cup c_j \cup T$ is assigned the terrain class \emph{step}. In addition, we compute the angle $\alpha_\text{i,j}$ between $c_i$ and $c_j$ and save it for $c_s$. Since most \emph{step} cells are detected several times, we collect several angles for each cell. $\alpha_\text{avg,s}$, the \emph{mean of circular quantities} of these angles describes the estimated step orientation in $c_s$. 

Second, we classify the remaining cells by their \emph{Level~2} height difference value $\bigtriangleup H$\footnote{These height difference values are subsampled and smoothed and thus cannot be directly transferred to occurring height differences in the terrain.}:
\begin{itemize}
	\item \emph{flat} if $\bigtriangleup H(c_i)\in [0\text{\,m},2*10^{-4}\text{\,m}]$,
	\item \emph{rough} if $\bigtriangleup H(c_i)\in [2*10^{-4}\text{\,m},0.05\text{\,m}]$,
	\item \emph{wall} if $\bigtriangleup H(c_i)\in [0.05\text{\,m},\infty]$, and
	\item \emph{unknown} if $\bigtriangleup H(c_i)$ is unknown.
\end{itemize}
The height difference intervals are tuned manually with respect to a maximum terrain height difference of 4\,cm which can be overcome by driving and a maximum terrain height difference of 30\,cm which can be overcome by stepping. The terrain class of a \emph{Level~3} map cell is generated from the respective four \emph{Level~2} cells by either choosing the terrain class with most members or, if this cannot be identified, the least difficult occurring terrain class.

\begin{figure}
\centering
\input{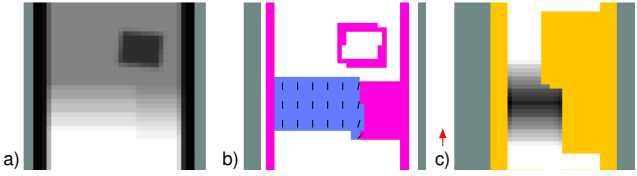}
\caption{\emph{Level~3}: a) height map, b) terrain class map (white = flat, \mbox{blue = stepping}, pink = wall, black lines = step orientations), c) robot area cost map for the orientation indicated by the red arrow.}
\label{fig:level3_maps}
\end{figure}

Another source for terrain class segmentation can be camera images as shown in~\cite{schilling2017geometric}. \cref{fig:level3_maps}\,a,b gives an example for a \emph{Level~3} height map and terrain class map.

The \emph{Level~3} robot representation $\vec{r}=\vec{r}_b$ only consists of the robot base pose. Individual feet positions are not considered but we assume that the feet are somewhere in a ground contact area around the robot $a_r$ (see~\cref{fig:representation_level_concept}). Hence, the robot is not able to perform  foot or foot pair movements in this representation. The whole robot is rather moved over the terrain and traverses different terrain classes with different costs. Path search neighbour poses can be found similar to the driving neighbours described for \emph{Level~1}. In this level of representation, the action resolution is 10\,cm and the robot may have 16 different orientations at each position. When moving over \emph{step} cells, a robot state is only feasible if the difference between the robot orientation and the step orientation of each \emph{step} cell $c_r$ is less than one discrete orientation step: $\text{abs}(\alpha_\text{avg,r} - r_\theta) < \frac{1}{16} \cdot 2 \pi$. Moreover, the robot is only allowed to move parallel and orthogonal to step orientations. These restrictions are required to enforce a behavior, which is induced by the robot kinematic in lower representation levels but not represented in \emph{Level~3} otherwise. 

Regarding cost generation, each cell $c_i$ is assigned a cost value $C_\text{c}(c_i)$ depending on its terrain class:
\begin{itemize}
	\item \emph{flat}: $C_\text{c}(c_i)$ = 1.0,
	\item \emph{rough}: $C_\text{c}(c_i)$ = 1.4,
	\item \emph{step}: $C_\text{c}(c_i)$ = 76.0 + 2.95 $\cdot \bigtriangleup H(c_i)$,
	\item \emph{wall}: $C_\text{c}(c_i)$ = $\infty$, and
	\item \emph{unknown}: $C_\text{c}(c_i)$ = nan.
\end{itemize}
The pose cost $C(\vec{r})$ does not combine individual ground contact and body cost but averages the cost values of all cells in $a_r$. The described terrain class specific cell costs are manually tuned by comparing the cost of \emph{Level~1} and \emph{Level~3} manoeuvres for the same set of basic scenarios, as mentioned in~\cref{sec:lvl2}. While constant values were sufficient for \emph{flat} and \emph{rough} cells, costs for stepping manoeuvres depend on the height difference to overcome. The presented computation method for \emph{step} cells is required to keep cost differences for these basic manoeuvres $\leq$\,5\%. A resulting robot area cost map can be seen in~\cref{fig:level3_maps}\,c.

\emph{Level~3} paths can be refined to \emph{Level~2} paths in the following way: As described for \emph{Level~2}, we generate a set $S$ of feasible robot base poses. In contrast to \emph{Level~2}, we do not only consider two successive poses but the whole path segment $\vec{r}_\text{3,s},..., \vec{r}_\text{3,g}$ that needs to be refined at once. The first and last robot pose $\vec{r}_\text{3,s}$ and $\vec{r}_\text{3,g}$ of this \emph{Level~3} path segment are transformed to a \emph{Level~2} start and goal pose and a local \emph{Level~2} planner, which is restricted to $S$, searches for a path between $\vec{r}_\text{2,s}$ and $\vec{r}_\text{2,g}$. If a \emph{Level~3} path needs to be refined to \emph{Level~1}, \emph{Level~2} is taken as an intermediate refinement step.


\subsection{Level Transition}

All three levels of representation are combined in a single planner, which chooses the lowest available level for each pose to provide the most detailed planning. Since planning in a low level of representation is slower, we provide \emph{Level~1} data only in a small area around the robot position which is sufficiently large to plan the next manoeuvres. \emph{Level~2} data is provided for a medium-sized region around the current robot position while \emph{Level~3} covers the whole map.

The planner checks for each manoeuvre (e.g., drive into one direction, do a step, ...) if both, start and goal pose of this manoeuvre, are part of the same level of representation. If the goal pose is not part of the start pose level of representation, the start pose is transformed to the next higher level of representation and the same manoeuvre is replanned in this level if it is still available in this level. Note that the transformation of the start pose to the next higher level of representation might induce costs. Due to different map resolutions, the robot might be shifted to fit into the next level map cell and discrete orientation. Due to increasing foot restrictions, feet  might be shifted to fit the next level robot representation (e.g., individual feet have to align within foot area pairs). We check each transformation for feasibility and generate costs from the occurring manoeuvre costs.


\subsection{Heuristic}

In our previous work, a combination of the Euclidean distance and the orientation difference was used as an admissible A* heuristic (\emph{Euclidean heuristic}). However, this heuristic does not consider the terrain which has large influence on the path costs. We propose a \emph{Level~3}-based heuristic which includes such terrain features (\emph{Dijkstra heuristic}).

After the goal pose $\vec{r}_\text{i,G}$ is set, it is transformed to \emph{Level~3}. We then start a one-to-any 3D Dijkstra search in \emph{Level~3} starting from $\vec{r}_\text{3,G}$. Hence, we get for each \emph{Level~3} pose a cost estimation to reach the goal pose. During path planning, we can estimate the costs from any robot pose to the goal by transforming it to \emph{Level~3} and get the respective cost value.

Note that the quality of this heuristic strongly depends on the quality of the ~\emph{Level~3} cost model in comparison to costs for the same manoeuvres in other levels of representation. Further note that we cannot prove that this heuristic always underestimates costs, which is necessary to prove admissibility for the generation of optimal paths. However, since we also utilize the suboptimal ARA* algorithm, we do not aim to generate optimal paths for a given problem. In fact, we focus on generating paths with satisfying quality in feasible time. The performance of this heuristic is evaluated in~\cref{sec:experiments}.


\subsection{Continuous Refinement}
\label{sec:continuous_refinement}

\begin{figure}
\centering
\includegraphics[width=\linewidth]{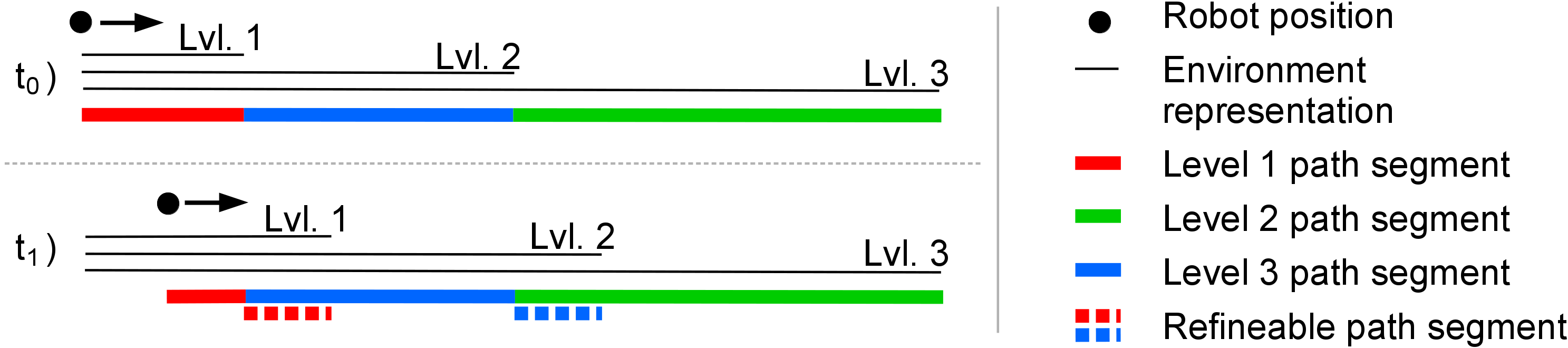}
\caption{As the robot moves along the path, the \emph{Level~1} and \emph{Level~2} representations move with it. Consequently, those path segments which are represented in a higher level and for which a more detailed representation becomes available, can be refined to this more detailed representation.}
\label{fig:continuous_refinement}
\end{figure}

As the robot moves along the initial path, the sensors provide new measurements and high-detailed environment representations are generated in the vicinity of the current robot position. We include these updated representations in the path by continuously refining the respective path segments, as shown in~\cref{fig:continuous_refinement}. If a cost difference \textgreater\,25\% between the original and the refined path segments indicates that the higher-level planning assessed a situation wrongly, we initiate a new planner run. With this approach, we can guarantee that path segments in the vicinity of the robot are always represented in ~\emph{Level~1} and thus, included steps can be expanded and the result can be executed by the controller.


\section{Experiments}
\label{sec:experiments}

We evaluate the proposed approach in two experiments. Both are done on one core of a 2.6\,GHz Intel \mbox{i7-6700HQ} processor using 16\,GB of memory. An additional video is available online\footnote{\url{https://www.ais.uni-bonn.de/videos/ICRA_2018_Klamt/}} which also contains a Gazebo experiment to demonstrate the continuous refinement strategy.

\begin{figure}
\centering
\input{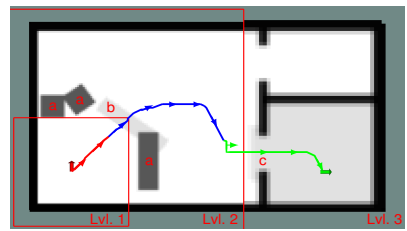}
\caption{Height map of the first experiment scenario. From its start position (red arrow), the robot needs to navigate between multiple objects (a), over a bar obstacle (b), step up to an elevated platform and through a door (c) to the goal pose (green arrow). The resulting path for $\mathcal{W} = 1.125$ and combined levels of representation is shown. \emph{Level~1} path segments = red, \emph{Level~2} segments = blue, \emph{Level~3} segments = green. Arrows show $r_\theta$.}
\label{fig:exp1_scenario}
\end{figure}

\begin{figure}
\centering
\input{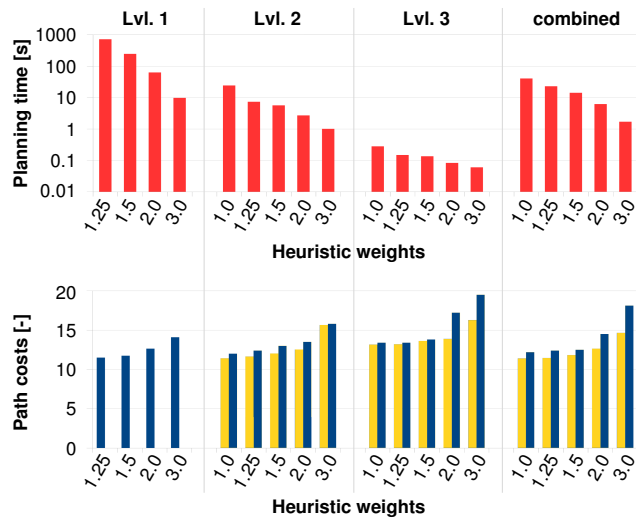}
\caption{Planning performance for different levels and different $\mathcal{W}$ for the first experiment. estimated costs = yellow, refined \emph{Level~1} costs = blue.}
\label{fig:exp1_result_charts}
\end{figure}

A first experiment evaluates the planning performance of the different levels of representation individually and combined, as shown in~\cref{fig:level_size_concept}. For this, we choose the \emph{Level~1} size to be 3$\times$3\,m. This is sufficiently large to plan the next robot manoeuvres in high detail, but still small enough to avoid long high-dimensional planning. The \emph{Level~2} size is chosen to be 9$\times$9\,m so that the \emph{Level~2} path segment is about twice as long as the \emph{Level~1} path segment. We utilized the \emph{Euclidean heuristic} to compare the results to our previous work. The height map and a resulting path are shown in~\cref{fig:exp1_scenario}. Since we use an ARA* algorithm which works with several heuristic weights $\mathcal{W}$, we evaluate the influence of these. \cref{fig:exp1_result_charts} shows the planner performance. 

It can be seen that planning on levels of representation \textgreater1 and with combined levels is faster by at least one order of magnitude compared to pure \emph{Level~1} planning. The \emph{Level~1} path for $\mathcal{W} = 1.0$ could not be computed due to memory limitations. We distinguish between the path costs in the respective levels of representation (estimated cost) and the costs each path carries when refined to \emph{Level~1}. Comparing the estimated costs to the refined \emph{Level~1} costs gives an assessment about the quality of cost generation in each level of representation. The comparison of the refined \emph{Level~1} costs to the original \emph{Level~1} costs indicates the quality of the resulting path. It can be seen that the estimated costs always underestimate the refined \emph{Level~1} costs. Especially for $\mathcal{W}\,\leq\,1.5$ the estimation is close with a difference $\leq$\,7.7\%. Furthermore, the results show that for $\mathcal{W}\,\leq\,1.5$ the refined \emph{Level~1} costs differ to the original \emph{Level~1} costs by $\leq$\,15\%.

\begin{figure}
\centering
\includegraphics[width=0.7\linewidth]{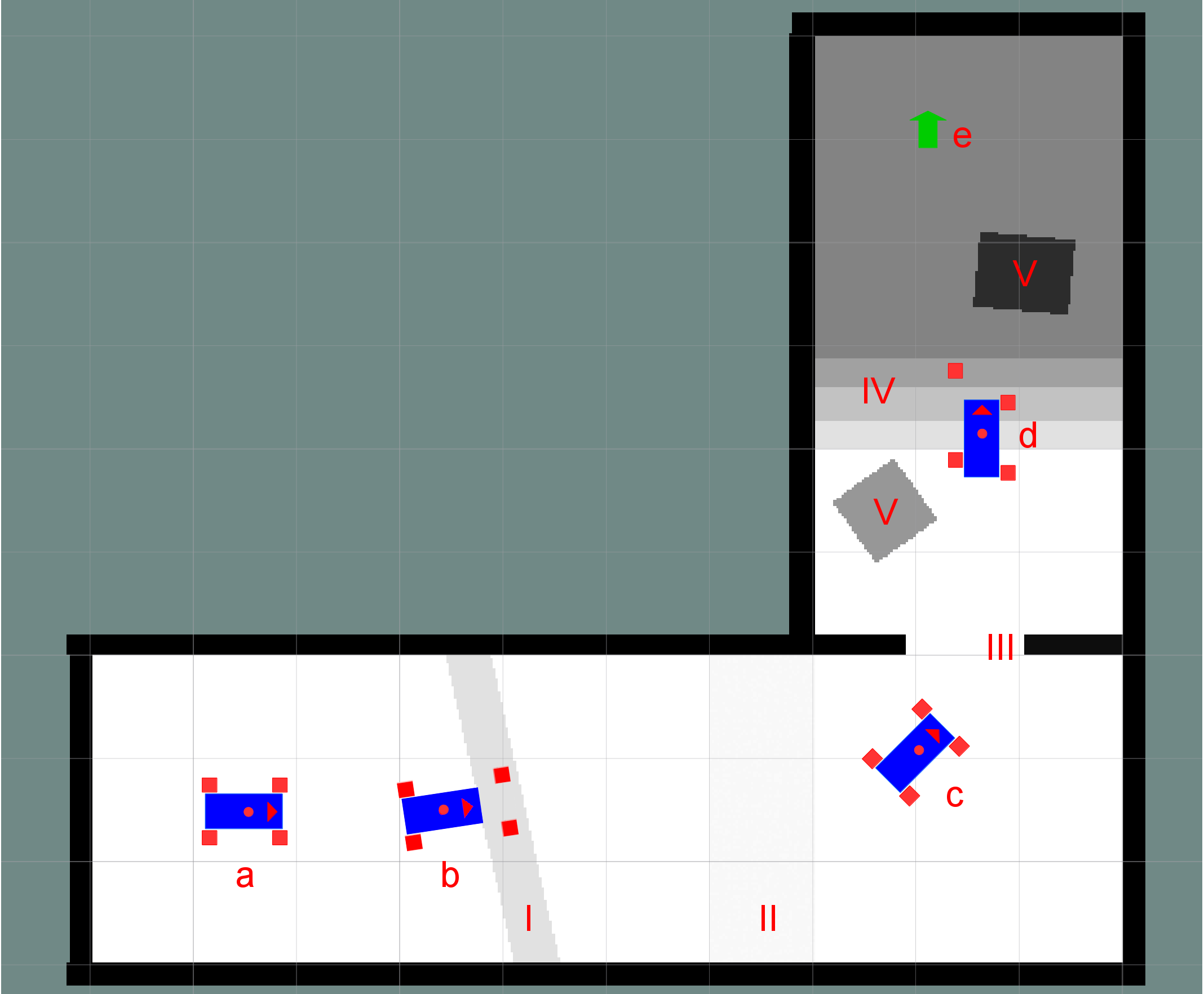}
\caption{Height map for the second experiment containing a bar obstacle (I), a rough area (II), a door (III), a flight of stairs (VI) and two obstacles (V). a - d are different starting poses for the planner, e is the goal pose.}
\label{fig:exp2_scenario_map}
\end{figure}

\begin{figure}
\centering
\input{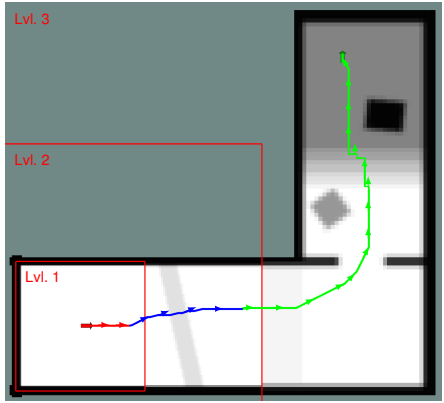}
\caption{Resulting path for planning with the \emph{Dijkstra heuristic} and combined levels with $\mathcal{W} = 1.25$.}
\label{fig:exp2_path_mixed}
\end{figure}

\begin{figure}
\centering
\input{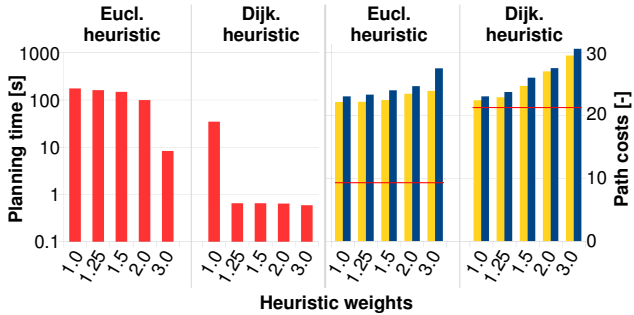}
\caption{Planning performance for combined levels of representation to compare the \emph{Euclidean heuristic} with the \emph{Dijkstra heuristic}. Red lines indicate the cost estimation for the path by each heuristic.}
\label{fig:exp2_level_compare_charts}
\end{figure}

In a second experiment, we compare the presented \emph{Dijkstra heuristic} to the \emph{Euclidean heuristic}. The scenario shown in~\cref{fig:exp2_scenario_map} is larger and more challenging, compared to the first scenario. The starting pose is pose \emph{a}. Planning is performed on combined levels of representation. A resulting path is shown in~\cref{fig:exp2_path_mixed}.  Planning times and resulting costs are shown in~\cref{fig:exp2_level_compare_charts}. Preprocessing the \emph{Dijkstra heuristic} took 0.52\,s of the presented planning times. It can be seen that the \emph{Dijkstra heuristic} further accelerates planning while the resulting costs stay comparable at least for $\mathcal{W}\leq1.5$. E.g., for \mbox{$\mathcal{W} = 1.25$}, planning is accelerated by more than two orders of magnitude while the refined path costs only differ by 3.3\%. Moreover, the resulting path illustrates how the robot aligns with the stairs and only moves parallel and orthogonal to them.

\begin{figure}
\input{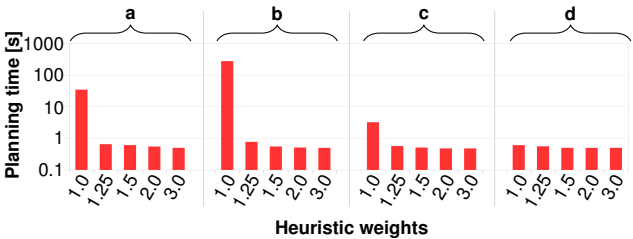}
\caption{Planning time for different starting poses (see~\cref{fig:exp2_scenario_map}) and different $\mathcal{W}$, using the \emph{Dijkstra heuristic}.}
\label{fig:exp2_position_compare_results}
\end{figure}

We finally compare the planner performance when started from different poses, as shown in~\cref{fig:exp2_scenario_map}. The results in~\cref{fig:exp2_position_compare_results} indicate that an important factor for the planner performance is the complexity of the planning within \emph{Level~1} but higher $\mathcal{W}$ lead to feasible performances in any case. 

\section{Conclusion}

In this paper, we presented a hybrid locomotion planning approach which is able to provide plans for large scenarios with high detailing in the vicinity of the robot. We achieve this by introducing three levels of representation with decreasing resolution and robot configuration dimensionality but increasing semantics of the situation. The most abstract level of representation can be used as a heuristic which poses a second acceleration strategy. Experiments show that the presented approach significantly accelerates planning while the result quality stays feasible and, hence, significantly larger scenarios can be handled in comparison to our previous work.



\bibliographystyle{IEEEtranN}
\bibliography{references}

\begin{thebibliography}{18}
\providecommand{\natexlab}[1]{#1}
\providecommand{\url}[1]{#1}
\csname url@samestyle\endcsname
\providecommand{\newblock}{\relax}
\providecommand{\bibinfo}[2]{#2}
\providecommand{\BIBentrySTDinterwordspacing}{\spaceskip=0pt\relax}
\providecommand{\BIBentryALTinterwordstretchfactor}{4}
\providecommand{\BIBentryALTinterwordspacing}{\spaceskip=\fontdimen2\font plus
\BIBentryALTinterwordstretchfactor\fontdimen3\font minus
  \fontdimen4\font\relax}
\providecommand{\BIBforeignlanguage}[2]{{%
\expandafter\ifx\csname l@#1\endcsname\relax
\typeout{** WARNING: IEEEtranN.bst: No hyphenation pattern has been}%
\typeout{** loaded for the language `#1'. Using the pattern for}%
\typeout{** the default language instead.}%
\else
\language=\csname l@#1\endcsname
\fi
#2}}
\providecommand{\BIBdecl}{\relax}
\BIBdecl

\bibitem[Klamt and Behnke(2017)]{klamtanytime}
T.~Klamt and S.~Behnke, ``Anytime hybrid driving-stepping locomotion
  planning,'' in \emph{IEEE/RSJ International Conference on Intelligent Robots
  and Systems (IROS)}, 2017.

\bibitem[Schwarz et~al.(2016)Schwarz, Rodehutskors, Schreiber, and
  Behnke]{Schwarz:ICRA2016}
M.~Schwarz, T.~Rodehutskors, M.~Schreiber, and S.~Behnke, ``Hybrid
  driving-stepping locomotion with the wheeled-legged robot {Momaro},'' in
  \emph{IEEE International Conference on Robotics and Automation (ICRA)}, 2016.

\bibitem[Bohlin(2001)]{bohlin2001path}
R.~Bohlin, ``Path planning in practice; lazy evaluation on a multi-resolution
  grid,'' in \emph{IEEE/RSJ International Conference on Intelligent Robots and
  Systems (IROS)}, 2001.

\bibitem[Kohrt et~al.(2012)Kohrt, Pipe, Kiely, Stamp, and
  Schiedermeier]{kohrt2012cell}
C.~Kohrt, A.~G. Pipe, J.~Kiely, R.~Stamp, and G.~Schiedermeier, ``A cell based
  {Voronoi} roadmap for motion planning of articulated robots using movement
  primitives,'' in \emph{IEEE International Conference on Robotics and
  Biomimetics (ROBIO)}, 2012.

\bibitem[Ziaei et~al.(2014)Ziaei, Oftadeh, and Mattila]{ziaei2014global}
Z.~Ziaei, R.~Oftadeh, and J.~Mattila, ``Global path planning with obstacle
  avoidance for omnidirectional mobile robot using overhead camera,'' in
  \emph{IEEE International Conference on Mechatronics and Automation (ICMA)},
  2014.

\bibitem[Howard and Kelly(2007)]{howard2007optimal}
T.~M. Howard and A.~Kelly, ``Optimal rough terrain trajectory generation for
  wheeled mobile robots,'' \emph{The International Journal of Robotics
  Research}, vol.~26, no.~2, pp. 141--166, 2007.

\bibitem[Brunner et~al.(2012)Brunner, Br{\"u}ggemann, and
  Schulz]{brunner2012motion}
M.~Brunner, B.~Br{\"u}ggemann, and D.~Schulz, ``Motion planning for actively
  reconfigurable mobile robots in search and rescue scenarios,'' in \emph{IEEE
  International Symposium on Safety, Security, and Rescue Robotics (SSRR)},
  2012.

\bibitem[Wermelinger et~al.(2016)Wermelinger, Fankhauser, Diethelm, Kr{\"u}si,
  Siegwart, and Hutter]{wermelinger2016navigation}
M.~Wermelinger, P.~Fankhauser, R.~Diethelm, P.~Kr{\"u}si, R.~Siegwart, and
  M.~Hutter, ``Navigation planning for legged robots in challenging terrain,''
  in \emph{IEEE/RSJ International Conference on Intelligent Robots and Systems
  (IROS)}, 2016.

\bibitem[Perrin et~al.(2016)Perrin, Ott, Englsberger, Stasse, Lamiraux, and
  Caldwell]{perrin2016continuous}
N.~Perrin, C.~Ott, J.~Englsberger, O.~Stasse, F.~Lamiraux, and D.~G. Caldwell,
  ``Continuous legged locomotion planning,'' \emph{IEEE Transactions on
  Robotics}, vol.~33, no.~1, pp. 234--239, 2016.

\bibitem[Behnke(2003)]{behnke2003local}
S.~Behnke, ``Local multiresolution path planning,'' in \emph{RoboCup 2003:
  Robot Soccer World Cup VII}.\hskip 1em plus 0.5em minus 0.4em\relax Springer,
  2003, pp. 332--343.

\bibitem[Gonz{\'a}lez-Sieira et~al.(2016)Gonz{\'a}lez-Sieira, Mucientes, and
  Bugar{\'\i}n]{gonzalez2016adaptive}
A.~Gonz{\'a}lez-Sieira, M.~Mucientes, and A.~Bugar{\'\i}n, ``An adaptive
  multi-resolution state lattice approach for motion planning with
  uncertainty,'' in \emph{Robot 2015: Second Iberian Robotics
  Conference}.\hskip 1em plus 0.5em minus 0.4em\relax Springer, 2016, pp.
  257--268.

\bibitem[Pivtoraiko and Kelly(2008)]{pivtoraiko2008differentially}
M.~Pivtoraiko and A.~Kelly, ``Differentially constrained motion replanning
  using state lattices with graduated fidelity,'' in \emph{IEEE/RSJ
  International Conference on Intelligent Robots and Systems (IROS)}, 2008.

\bibitem[Gochev et~al.(2011)Gochev, Cohen, Butzke, Safonova, and
  Likhachev]{gochev2011path}
K.~Gochev, B.~Cohen, J.~Butzke, A.~Safonova, and M.~Likhachev, ``Path planning
  with adaptive dimensionality,'' in \emph{Fourth Annual Symposium on
  Combinatorial Search}, 2011.

\bibitem[Zhang et~al.(2012)Zhang, Butzke, and Likhachev]{zhang2012combining}
H.~Zhang, J.~Butzke, and M.~Likhachev, ``Combining global and local planning
  with guarantees on completeness,'' in \emph{IEEE/RSJ International Conference
  on Robotics and Automation (ICRA)}, 2012.

\bibitem[Petereit et~al.(2013)Petereit, Emter, and Frey]{petereit2013mobile}
J.~Petereit, T.~Emter, and C.~W. Frey, ``Mobile robot motion planning in
  multi-resolution lattices with hybrid dimensionality,'' \emph{IFAC
  Proceedings Volumes}, vol.~46, no.~10, pp. 158--163, 2013.

\bibitem[Droeschel et~al.(2017)Droeschel, Schwarz, and
  Behnke]{droeschel2017continuous}
D.~Droeschel, M.~Schwarz, and S.~Behnke, ``Continuous mapping and localization
  for autonomous navigation in rough terrain using a {3D} laser scanner,''
  \emph{Robotics and Autonomous Systems}, vol.~88, pp. 104--115, 2017.

\bibitem[Likhachev et~al.(2003)Likhachev, Gordon, and Thrun]{likhachev2003ara}
M.~Likhachev, G.~J. Gordon, and S.~Thrun, ``{ARA}*: Anytime {A*} with provable
  bounds on sub-optimality,'' in \emph{Conference on Neural Information
  Processing Systems (NIPS)}, 2003.

\bibitem[Schilling et~al.(2017)Schilling, Chen, Folkesson, and
  Jensfeld]{schilling2017geometric}
F.~Schilling, X.~Chen, J.~Folkesson, and P.~Jensfeld, ``Geometric and visual
  terrain classification for autonomous mobile navigation,'' in \emph{IEEE/RSJ
  International Conference on Intelligent Robots and Systems (IROS)}, 2017.

\end{thebibliography}

\addtolength{\textheight}{-12cm}   





\end{document}